\documentclass[conference]{IEEEtran}
\IEEEoverridecommandlockouts
\usepackage{graphicx}
\usepackage{cite}
\usepackage{float}
\usepackage{amsmath,amssymb,amsfonts}
\usepackage{algorithmic}
\usepackage{textcomp}
\usepackage{xcolor}
\usepackage{balance}
\usepackage{soul}
\def\BibTeX{{\rm B\kern-.05em{\sc i\kern-.025em b}\kern-.08em
    T\kern-.1667em\lower.7ex\hbox{E}\kern-.125emX}}
\begin{document}

\title{Validating a Cortisol-Inspired Framework for Human-Robot Interaction with a Replication of the Still Face Paradigm}
%


\author{\IEEEauthorblockN{Sara Mongile}
\IEEEauthorblockA{\textit{DIBRIS, University of Genoa} \\
\textit{\& CONTACT Unit,}\\
\textit{Italian Institute of Technology}\\
Genoa, Italy \\
sara.mongile@iit.it}
\and
\IEEEauthorblockN{Ana Tanevska}
\IEEEauthorblockA{\textit{CONTACT Unit} \\
\textit{Italian Institute of Technology}\\
Genoa, Italy \\
ana.tanevska@iit.it}
\and
\IEEEauthorblockN{Francesco Rea}
\IEEEauthorblockA{\textit{RBCS Department} \\
\textit{Italian Institute of Technology}\\
Genoa, Italy \\
francesco.rea@iit.it}
\and
\IEEEauthorblockN{Alessandra Sciutti}
\IEEEauthorblockA{\textit{CONTACT Unit} \\
\textit{Italian Institute of Technology}\\
Genoa, Italy \\
alessandra.sciutti@iit.it}
}

\maketitle

\begin{abstract}


When interacting with others in our everyday life, we prefer the company of those who share with us the same desire of closeness and intimacy (or lack thereof), since this determines if our interaction will be more o less pleasant. This sort of compatibility can be inferred by our innate attachment style. The attachment style represents our characteristic way of thinking, feeling and behaving in close relationship, and other than behaviourally, it can also affect us biologically via our hormonal dynamics. When we are looking how to enrich human-robot interaction (HRI), one potential solution could be enabling robots to understand their partners' attachment style, which could then improve the perception of their partners and help them behave in an adaptive manner during the interaction. We propose to use the relationship between the attachment style and the cortisol hormone, to endow the humanoid robot iCub with an internal cortisol inspired framework that allows it to infer participant's attachment style by the effect of the interaction on its cortisol levels (referred to as \textit{R-cortisol}). In this work, we present our cognitive framework and its validation during the replication of a well-known paradigm on hormonal modulation in human-human interaction (HHI) - the Still Face paradigm.

\end{abstract}


\section{Introduction}

People have a natural predisposition to interact in an adaptive manner with others, by instinctively changing our actions, tones and speech according to the perceived needs of our peers. However, we also have a preference for interacting with partners who share with us the same desire of closeness and intimacy (or lack thereof), since this leads to an interaction that is a pleasure and not a stressor according to the similarity theory of attachment \cite{holmes2009adult}. This perceived sort of "compatibility" can be related to our innate \textit{attachment style}. The attachment style is a person’s characteristic way of forming relationships and modulating behavior (i.e., partner's perception, or ways to give or seek support). It develops around the first year of life, strongly influenced by the relation with the caregivers \cite{ainsworth2015patterns}. In children, the attachment styles can be classified into four prototypes: secure attachment and three insecure styles - ambivalent-anxious, avoidant and disorganized \cite{ainsworth2015patterns}. 

Attachment style affects also our perception of our partners, as well as the regulation of our emotions, which is reflected on a biological level directly in different hormones dynamics, such as in the hypothalamic–pituitary–adrenal (HPA) axis activation \cite{kidd2013adult, gunnar2002social}. People with different attachment styles experience differently the same social situations (i.e., some people may perceive a stimulus as a pleasing one, while to others the same one would be a stressor). This sensitivity results in an increase or decrease of the cortisol levels after being exposed to the same situation; in particular people with any insecure attachment style report higher cortisol levels than secure people after have experienced the same stressor \cite{gander2015attachment, spangler1998emotional}. Moreover, interaction between strangers that have the same attachment style leads to lower cortisol levels in both compared to interaction with strangers with a mismatch in attachment styles \cite{ketay2017attachment}. This is evidences the importance of interacting with partners whose perception of closeness and space is similar to ours, and whose behavior is not perceived as a stressor. 

This can be seen also as a key point in human-robot interaction (HRI), since to become effective partners robots need to adapt their behavior according to the human partner's need and affective states \cite{sciutti2018humanizing}. To reach this goal, our proposal is to endow our robots with an internal cortisol framework, in which the robot's cortisol levels (\textit{R-cortisol}) will change as a result of its own attachment style and the behavioural manner of its partner.

In particular, we wish to endow the humanoid robot iCub \cite{metta2008icub} with an internal motivation drive taking inspiration from cortisol as a modulator. Through the cortisol-inspired framework iCub will be able to understand the human partner's attachment styles by the effect the interaction will have on its R-cortisol levels \cite{mongile2022robots}. The perceived attachment style will then be used by the robot to adapt its behavior accordingly.

We start from existing hormonal framework models in HRI \cite{tanevska2020socially,hiolle2012eliciting,avila2005hormonal} and from different studies in HHI focusing on the relationship between cortisol and attachment style \cite{kidd2011examining,borelli2014dismissing}, both in children and in adults. From there, we design two robotic behavioral profiles portraying two attachment styles (an avoidant and an anxious attachment style). We assume that, as it happens during human-human-interaction, a mismatch during human-robot interaction in the expressed attachment style between a participant and a robot will cause higher R-cortisol levels in the robot than a match in attachment style \cite{barbosa2020emerging,leerkes2012infant,ketay2017attachment}. We validate our cortisol-inspired framework with two robot profiles in a small validation study with the humanoid robot iCub, before testing the cortisol-inspired framework in a real interaction with participants. The validation study which is the focus of this paper is a loose replication of the Still Face \cite{tronick1978infant} and Still Face+Touch paradigms \cite{stack1992adult}, which are two well-known paradigms in hormonal studies in HHI used between mothers and infants to elicit cortisol increase in infants. They consist in three brief episodes of interaction structured in A-B-A sequences, where during the B phase the stress is induced through the manipulation of mother's behavior.

The rest of the paper is organized as follows: Section II presents the materials and methods used in our research, with Subsections A. and B. presenting the cognitive framework with hormonal modulation with its components and their functionalities, and the designed robot profiles. This is followed by Subsections C. and D. which present the Still Face Paradigm and the validation study in which we tested the performance of our framework in both robot profiles with naive participants. Finally, in Sections III and IV we present the results from our studies and we discuss our plans for future work.

\section{Methodology}

\begin{figure}[h]
\includegraphics[width=\columnwidth]{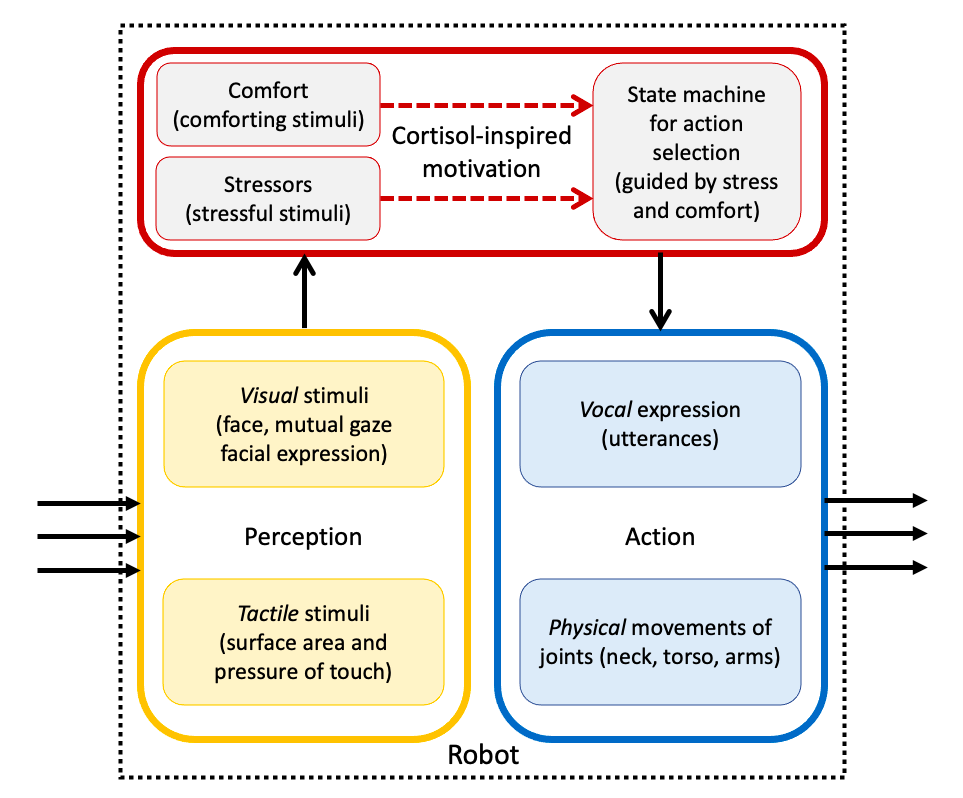}
\caption{Three main components of the framework - Perception, Action and Motivation Module. Perception component receives and processes Visual stimuli and Tactile stimuli. Visual stimuli consist of detection of face, facial expression, and presence of mutual gaze. Tactile stimuli consist of surface area of the touch and intensity or pressure. Action component sends the Vocal and Physical expressions of the robot. The vocal expression represents the robot's utterances. The physical expression are the movements of the robot's body parts - neck, torso and arms. The motivation component analyzes the received data from the Perception component and sorts it in two categories - Comforting stimuli and Stressful stimuli. These two categories are analyzed in the cortisol-inspired motivation module and are sent to the State Machine that guides the action selection process for the robot, which is then connected to the Action component.}
\label{fig:cognFramework}
\end{figure}

In our research, we have implemented a cognitive framework for our robots to study in a more structured manner how the motivational mechanism rooted in cortisol changes during HRI \cite{mongile2022robots}. Our framework aims to provide the robots with the primary supportive functionalities necessary for the HRI studies. It consists of the following modules and their functionalities (as illustrated in Figure \ref{fig:cognFramework}): a \textit{perception} module processing tactile and visual stimuli; an \textit{action} module, responsible for the robots' movements; and a \textit{motivation} module, containing the cortisol-inspired internal motivation. To validate the full functionality of the framework according to findings in literature, we then designed two robot profiles that interact with the framework in different ways, mimicking the attachment styles of children.

The aim of this work was to validate our cortisol-inspired framework and test both robot profiles by conducting a replication of the \textit{Still Face Paradigm}. This is a well-known paradigm in child-caretaker interaction, used both to elicit a cortisol increase, as well as to evaluate children's attachment style. With our validation study, we wish to evaluate if our framework and the designed profiles for the robot function as we expect them to. 

\subsection{The Cognitive Framework}

The \textit{perception} module processes stimuli from two sensor groups of the robot: \textit{visual stimuli} and \textit{tactile stimuli}. The visual stimuli, provided as stream of images, are received through the robot's camera situated in its eyes (although usually it is sufficient to use the images only from one of the two cameras, in this case the left eye). They are analyzed for detecting the presence of a person's face, extracting the facial features of the person using open-source library OpenFace \cite{baltruvsaitis2016openface}. The data from the OpenFace library is then processed for obtaining the most salient action units (AU) \cite{ekman1978facial} from the detected facial features - lowering/raising eyebrows, crinkling of nose and cheeks, and smiling/frowning, as well as for detecting a potential mutual gaze. The tactile stimuli are collected from a skin sensor patches covering the robot's arms and torso \cite{cannata2008embedded}, which carry the information about the size of the area that is being touched (expressed in number of \emph{taxels} - tactile elements) and the average pressure of the touch.

The \textit{action} module performs a finite set of actions by controlling the specific body parts in joint space control of the robot's neck, torso and arms. The first experimental studies will focus on loosely replicating the results from human child-caretaker studies, where the robot will be in the role of a toddler. The robot's childlike role is essential for the replication of the human child-caretaker studies, and this is further assisted by iCub's childlike appearance; as such, the actions performed by the robot are limited to a list of simpler behaviours: a) turning the torso towards the participants, b) stretching its arms with open hands towards them with a smiling face to seek contact and c) requesting attention by calling out vocally to them. Although these are fairly simplistic behaviours, in the context of HRI they have proven effective in as shown in our previous studies \cite{tanevska2020socially,tanevska2019cognitive}.

The \textit{motivation} module contains the implementation of our proposed cortisol-inspired internal framework, loosely inspired by previous studies in HRI with hormonally-based internal motivations for the robot \cite{tanevska2020socially,hiolle2012eliciting,avila2005hormonal}. The robot’s internal state changes as a function of the perceived change in the person’s affective state, or the actions performed by the human partner. Our framework processes the visual and tactile stimuli received by the human to update the stress and comfort levels, which in turn directly influence the R-cortisol level. The robot relies on its R-cortisol levels and its current behavioral state to guide its behaviour and select its next action. In this phase, taking inspiration from literature on human-human (and more specifically, child-caretaker) interaction, we designed two robot profiles mimicking two different child attachment styles.

\subsection{The Robot Profiles}
Taking inspiration from studies of child-caretaker interaction, we designed two robots profile with different attachment styles: one high in the anxiety dimension (\textit{Anxious} profile), and another high in avoidance dimension (\textit{Avoidant} profile). These profiles are designed to have different sensitivity to the human stimulation which is reflected in their R-cortisol patterns and their different behavior after being exposed to the same stimuli. The anxious profile perceives physical contact as a comfort, while its absence is perceived as a stressor; this instead is the opposite for the avoidant profile which perceives the constant touch as an intrusive and stressful signal. Moreover, the avoidant profile perceives the absence of interaction as a normal situation and not as a stressor, and this situation does not elicit any R-cortisol reactions like it does for the anxious profile. These profiles have also different R-cortisol dynamics: the avoidant profile has a higher R-cortisol threshold and faster recovery than the anxious profile, while the anxious profile is more reactive and has a longer recovery than avoidant profile.  


\subsection{The Still Face Paradigm}

Our validation study consists of replicating a well-known paradigm in hormonal studies in HHI between mothers and infants: the Still-Face Paradigm (SF) \cite{tronick1978infant}. The paradigm typically consists of three brief episodes, structured in an A-B-A sequence \cite{provenzi2016infants}: the first “A” corresponds to the \textit{Play} episode, where mothers and infants interact in a normal dyadic interaction setting; the “B” corresponds to the \textit{Still-Face} episode, in which mothers are asked to become completely unresponsive, not provide any touch or feedback, and maintain a neutral facial expression; and finally the second “A” is the \textit{Reunion} episode, where mothers and infants restart their normal interaction, which presents the context of socio-emotional stress recovery. The \textit{Still Face} phase induces a socio-emotional stress through the experimental manipulation of the mother's responsiveness and availability to interact, which elicits a cortisol increase in children.

Other studies have proposed a modification of the SF paradigm to include the maternal touch during the SF episode, creating a new paradigm - Still Face + Touch (SF+T). They examined the differential effects of touch versus no-touch conditions on infant behaviors \cite{stack1992adult}; the comparison shows lower cortisol levels in infants in SF+T condition, pointing to a more attenuate response. Moreover, cortisol decreased at recovery for the SF+T condition and it markedly increased for SF condition indicating continued stress during reunion \cite{feldman2010touch}. Pairing this knowledge with other findings in literature for children \cite{barbosa2020emerging,gander2015attachment} and adults \cite{simpson1992support}, where anxious individuals require more tactile and intense interaction, whereas avoidant ones prefer to receive less stimuli, for our validation study we assume that this holds true for the anxious profile, while the avoidant one perceives as a stressor the Still Face+Touch paradigm instead of the Still Face. 

\subsection{Validation Study with iCub}

After the completed implementation of the cortisol-inspired framework, we performed a validation study to verify the different attachment profiles and the R-cortisol model. 

\subsubsection{Experimental design}

To facilitate the conducting of repeated trials during the validation study, we prerecorded six sets of human stimuli, covering the 2 paradigms (SF and SF+T) and 3 human behavioral profiles - control, avoidant and anxious. We used these stimuli sets during a \textit{pre-test phase} in the process of fine-tuning the parameters of the cortisol-based motivation module, with the goal of obtaining the same results present in HHI literature \cite{mongile2022robots}. 

Following that, we ran a validation study with naive users and the humanoid robot iCub to assess whether the framework could successfully guide a real interaction. However, before implementing the robot's final behaviors in the action module, we decided to present the robot as static, using solely its functionalities to record stimuli during the interaction. We proceeded in this way since we wished to first understand if the planned actions of the robot would be compatible with the interaction setup, as well as see if the proposed perceptual modalities (touch and facial expressions) would be sufficient and represent well how participants would interact with a robot; or whether participants may prefer to use other modalities instead, like for example vocal interaction or playing with toys.
For these reasons, during the experiment the robot's function was to record visual and tactile stimuli (i.e., facial expression, presence of face, touch), and the experimenter narrated how the robot would have behaved in that situation for the two robot profiles (the anxious and the avoidant). The narration was reliable and consistent, corresponding always to the robot's profile and R-cortisol levels and the participants' actions (e.g., the avoidant robot with a high R-cortisol level always pulled away from a participant that tried to touch it). 

The experiment was conducted with six naive users (all female, mean age 26 $\pm$ 2.36), where three of them interacted with a narrated avoidant robot, and the other three with a narrated anxious robot. 

\begin{figure*}[htb]
\centering
\includegraphics[width=0.8\textwidth]{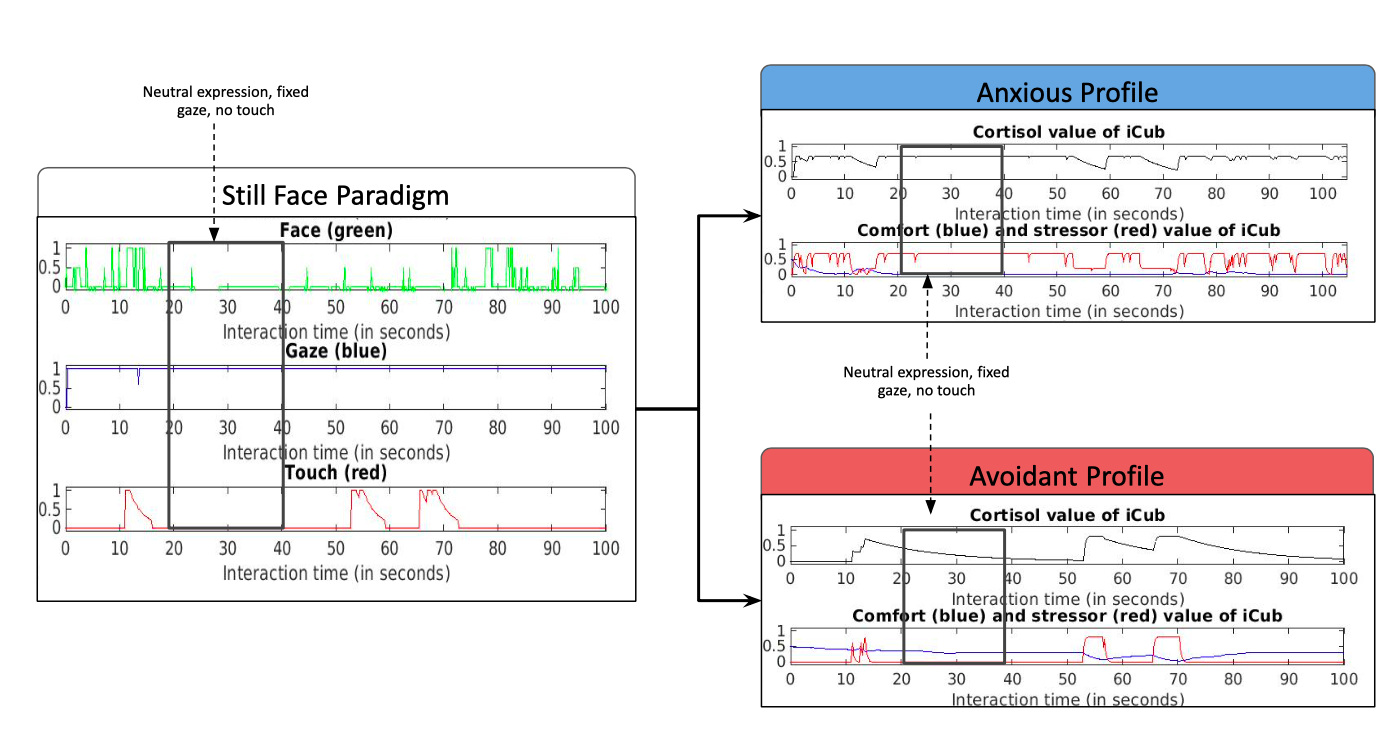}
\caption{Comparison between anxious and avoidant R-cortisol trends after being exposed to the same set stimuli. The robot profiles are designed to be sensitive to different stimuli, reflected in the R-cortisol reactions. The Still Face paradigm elicits higher R-cortisol reaction in the anxious profile than sin the avoidant one.}
\label{fig:comparison}
\end{figure*}

\subsubsection{Protocol}
In the laboratory iCub was positioned in front of a table. The participants were offered a chair from the other side of the table and they were asked to interact with the robot. The interaction scenario placed iCub in the role of a toddler, while the participants were tasked as the iCub's caretaker, for two brief sessions of two minutes each. In the two sessions, they had to replicate the Still Face and the Still Face+Touch paradigm (the order of the two was not the same for all participants).
In both paradigms, for the first 20 seconds participants were free to interact as they wish, then they were instructed to begin with the paradigm phase and maintain a fixed behavior for another 20 seconds, after which they could return to interact freely with the robot. Each interaction lasted 2 minutes (20'' free play, 20'' paradigm (SF or SF+T), 20'' reunion, 60'' free play).


During all of the experiment, the robot was in an active position (with its motors and facial LEDs turned on), with a smiling face and straight posture, but not moving or reactive. The experimenter narrated the robot's behavior for the two robot profiles (the anxious and the avoidant), giving different information considering participant's behavior. For example, if the participant was ignoring the robot in anxious profile, the narrated information given was "The robot stretches its arms towards you.", whereas if the robot was in the avoidant profile the information was "The robot looks away and tries to avoid you.".

\subsubsection{Data Analysis}

During the experiment we collected the data from the perception module: the participants' facial expression, the presence of a mutual gaze, and the presence of touch. We then used this stimuli to run a set of simulations to quantify the R-cortisol values for the robot during all interactions for the anxious and the avoidant profiles.

\section{Results}

The recorded data were analysed in order to verify how the cortisol-inspired framework reacts with different kind of stimuli combinations. 


Since with the Omnibus test we did not find any significant effect of the "narrated robot behavior" on the R-cortisol neither in the Still Face ($\chi_2 = 1.67, \rho$ = 0.19) nor in Still Face + Touch ($\chi_2 = 0.08, \rho = 0.77$) we did not consider the narration as a factor in all subsequent analysis. 
In the following we compare how the stimuli collected during the interactions would impact on the R-cortisol of the two different robot profiles.


In Figure \ref{fig:comparison} we report stimuli from the Still Face paradigm recorded during an interaction 
and the R-cortisol reactions in both profiles. The avoidant profile showed low R-cortisol levels for the entire interaction, whereas the same stimuli used in a simulation of an anxious robot induced higher R-cortisol levels for all the interaction. This was due to the different sensitivity of the two profiles: the absence of touch and smile are not a stressor for the avoidant profile, while they are for the anxious profile. Moreover, we can observe R-cortisol increase in the avoidant profile after the robot has been touched, while in the same point the R-cortisol in the anxious profile decreases.

\begin{figure}[h]
\centering
\includegraphics[width=0.8\columnwidth]{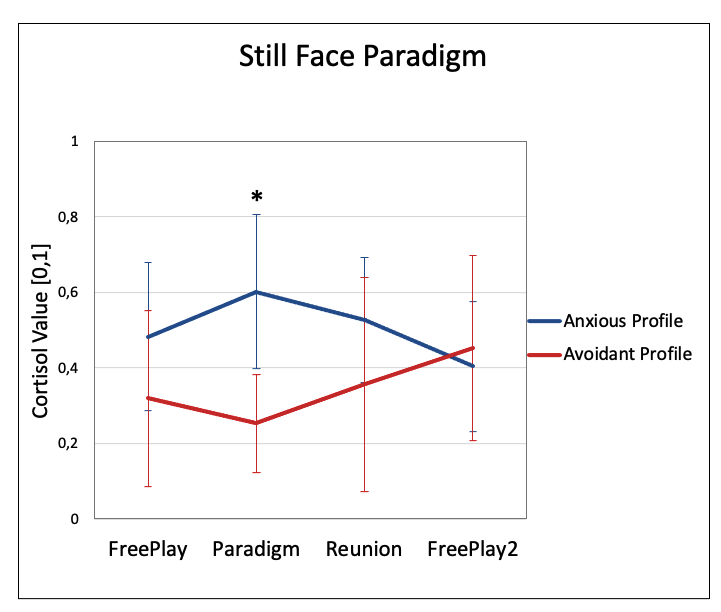}
\caption{Comparison of R-cortisol trends during the Still Face paradigm between avoidant and anxious profile. The paradigm induces significantly (*) higher R-cortisol levels in anxious profile than avoidant profile.}
\label{fig:cortisolTrendsSF}
\end{figure}

\begin{figure}[h]
\centering
\includegraphics[width=0.8\columnwidth]{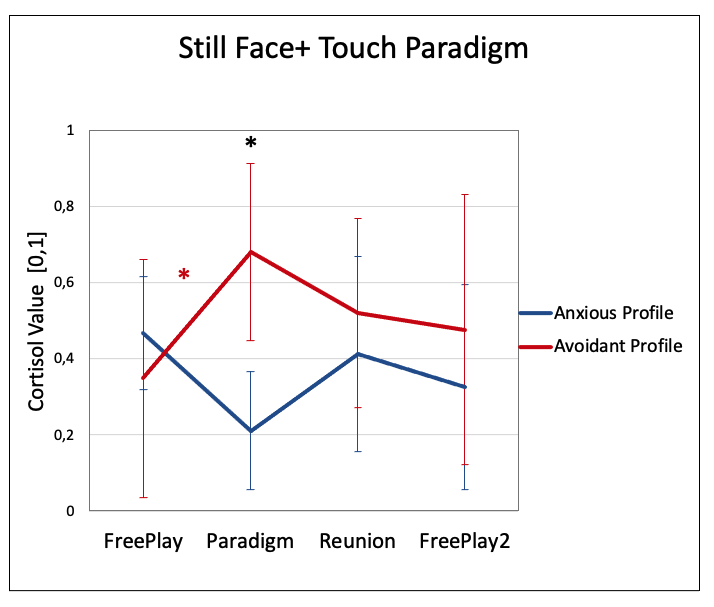}
\caption{Comparison of R-cortisol trends during the Still Face + Touch paradigm between avoidant and anxious profile. The paradigm induces significantly (*) lower R-cortisol levels in anxious profile than avoidant profile. In the avoidant profile there is also a significant difference between the average R-cortisol levels during the free play and during the paradigm. 
}
\label{fig:cortisolTrendsSFT}
\end{figure}

We evaluate the effects of the paradigms calculating the average of R-cortisol during the different phases of the interaction: 20'' of free play (FP), 20'' of paradigm (P), 20'' of reunion (R) and for all the rest of interaction (60'', free play - FP2). We compare R-cortisol values in the anxious and in avoidant profiles to see how the same stimuli affect the R-cortisol dynamics in each phase. During the Still Face (Figure \ref{fig:cortisolTrendsSF}), the anxious profile shows higher R-cortisol levels than the avoidant profile. In particular, the paradigm shows a tendency to increase R-cortisol in the anxious profile.
Instead, the avoidant profile shows higher R-cortisol levels during the Still Face + Touch (as shown in Figure \ref{fig:cortisolTrendsSFT}.

We run a statistical analysis using the mixed-effects model with R-cortisol values as dependent variable, robot profiles and phase and their interactions as independent factors and subject as random effect.

This analysis showed a significant effect for the robot profile during the paradigm phase, both during the Still Face Paradigm ($\beta = -0.35, z= -3.18, p-value= 0.001$) and the Still Face+Touch ($\beta = -0.46, z= -3.49, p-value= 0.000$). In particular during the Still Face the average R-cortisol level in the anxious profile is significantly higher than in avoidant; instead during the Still Face+Touch paradigm the average R-cortisol level in the avoidant profile is significantly higher than in anxious. Moreover, in the avoidant profile R-cortisol average is significant higher during the Still Face+Touch paradigm than during the free play ($\beta = -0.33, z= -2.47, p-value= 0.014$). These preliminary results are coherent with Feldman study \cite{feldman2010touch} and with our assumption that constant touch induces higher R-cortisol levels in the avoidant profile.

Then, we assess R-cortisol reactions during the entire interaction considering the relationship between the average R-cortisol value and the stimuli received. In particular, for each interaction we evaluate the participant's behavior based on the percentage of time spent touching the robot ("percent of touch") 
and percentage of time spent smiling during the interaction ("percent of smile"). These percentages are calculated using the total number of "perceptual" frames collected by the robot during the interaction, and evaluating in how many there is registered touch or smile by the participant. We define the interactions where the sum of percent of touch+percent of smile is higher than 35\% "interactive", while the others are defined "not interactive".

\begin{figure}[h]
\centering
\includegraphics[width=0.8\columnwidth]{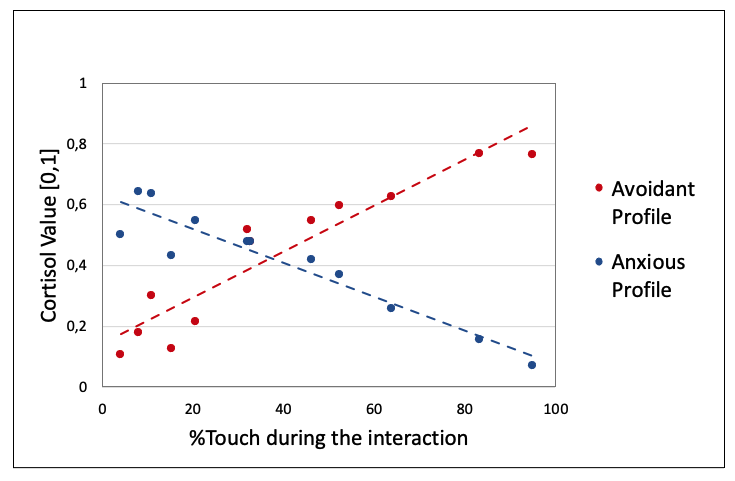}
\caption{The effect of touch on the R-cortisol average value during the interaction in anxious and avoidant profile. The avoidant profile reacts with higher R-cortisol level when the percent of touch is higher, while the anxious profile perceives as more stressful - higher R-cortisol levels- the absence of touch.}
\label{fig:touchEffect}
\end{figure}

We perform an analysis to see how the percent of touch during the entire interaction affects the average R-cortisol levels in anxious and avoidant profile. Figure \ref{fig:touchEffect}, where each point corresponds to a participant, shows that an higher percent of touch induces higher R-cortisol levels in avoidant and low R-cortisol level in anxious. 

          %

Finally, we assess if a "match" between the type of interaction (i.e., "interactive" and "not interactive) and the robot profiles induces a lower R-cortisol level, while a "mismatch" induces higher R-cortisol levels. 
We calculate the percentage of time during all the interaction in which the R-cortisol is over threshold (determined as the half of the cortisol maximum value for each profile).  Then, we consider a "match" the association between anxious profile and "interactive" interaction, and between avoidant profile and "not interactive" interaction. The opposite coupling \textit{vice versa} is considered a "mismatch".

\begin{figure}[h]
\centering
\includegraphics[width=0.8\columnwidth]{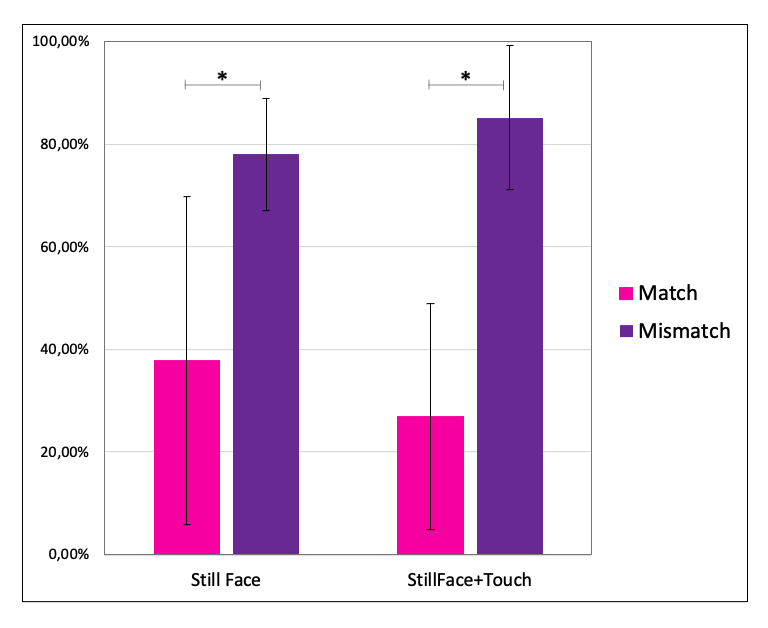}
\caption{Comparison between the percentage of R-cortisol over threshold during a match and a mismatch between participants' interaction style and the robot's attachment style in interaction. 
R-cortisol is significantly more often over threshold 
when there is a mismatch, both during the Still Face and during the Still Face + Touch.
}
\label{fig:match}
\end{figure}

The results (Figure \ref{fig:match}) show that 
R-cortisol is more often high, during the entire interaction, when there is a mismatch between the robot's profile and the person's style of interaction. 
A Wilcoxon signed rank test confirmed a significant difference 
between the match and the mismatch, both during the Still Face Paradigm ($z=2.2014, p-value=0.0277$) and during the Still Face+ Touch Paradigm ($z=2.2014, p-value=0.0277$). This preliminary finding is the first step toward our central questions: can we use robot's R-cortisol to understand which kind of person was interacting with the robot?. 

\section{Discussion and future work}

In our research, we aim to endow the humanoid robot iCub with an internal cortisol-inspired motivation that will allow it to infer its partner's attachment style from the effects of the interaction on its R-cortisol dynamics. Through this hormonal motivation we wish to improve the robot's perception of the partner's affective state. Moreover, this knowledge will drive the robot to adapt its behavior according to the partner's perceived attachment style. The first step for achieving our goal was the evaluation of our framework in a validation study with naive participants. The validation study consisted of a replication of the Still Face and the Still Face+Touch Paradigm, a well-known paradigm in human-human interaction used to elicit cortisol increase in toddlers. 

We designed two robot profiles inspired by the children attachment styles of avoidant and anxious. They behaved in different ways and were sensitive to different stimuli. These characteristics were then reflected in different R-cortisol reactions after experiencing the same situation. 
This preliminary study allowed us to verify how different stimuli combinations affect R-cortisol in the two robot profiles
From our first findings, we can confirm that our framework indeed mimics the hormonal dynamics of human-human interaction when modulated by specific social stimuli.
Even more importantly, this study gave us a first overview of the modalities used by naive users to interact. During the data collection, we observed that some participants preferred to speak with the robot much more than touching it. Since this is a way to establish an interaction, we will build on our existing perception module and integrate a novel audio module able to detect the occurrence of speech during the interaction. 

The analysis of the collected data show that, 
considering the average of R-cortisol during the interaction 
keeping in mind the robot profile, we get a glimpse of the kind of interaction the robot had. In particular, an interaction full of touches and smiles leads to high R-cortisol levels in the avoidant robot and significantly lower ones in the anxious robot. This represents a first small step towards verifying if R-cortisol can be a good one-dimensional parameter for understanding the partner's attachment style.

The next steps in our research will involve testing the framework in a comparative study, where the participants will entertain a full free-form interaction with the iCub, who will exhibit an avoidant or an anxious profile.


The aim is to assess how participants perceive the two robot's profiles and discover if there is a dedicated attachment style in human-robot interaction. (

This paper covered the first steps of our research in the direction of more congenial and pleasant human-robot interaction. By endowing robots with the capability to infer a person's attachment style and have a mirror of it in their own cortisol framework, we hope to improve their ability to adapt their behaviours, leading to a more natural and adaptive interaction between human and robots.




\section*{Acknowledgment}
The authors express their thanks to Dr. Joshua Zonca for his support and availability during the statistical data analysis.

This work has been supported by a Starting Grant from the European Research Council (ERC) under the European Union’s H2020 research and innovation programme. G.A. No 804388, wHiSPER.

\balance

\bibliographystyle{IEEEtran}
\bibliography{conference_101719}

\end{document}